\documentclass[10pt,twocolumn,letterpaper]{article}

\usepackage{cvpr}              %

\usepackage[pagebackref,breaklinks,colorlinks]{hyperref}

\usepackage{graphicx}
\usepackage{amsmath}
\usepackage{amssymb}
\usepackage{booktabs}
\usepackage{multirow, makecell}
\usepackage{graphicx}
\usepackage{adjustbox}
\usepackage[accsupp]{axessibility}

\usepackage{algorithm}
\usepackage{algpseudocode}

\usepackage[capitalize]{cleveref}
\crefname{section}{Sec.}{Secs.}
\Crefname{section}{Section}{Sections}
\Crefname{table}{Table}{Tables}
\crefname{table}{Tab.}{Tabs.}

\def\confName{CVPR}
\def\confYear{2024}

\usepackage{color}
\definecolor{alizarin}{rgb}{0.82, 0.1, 0.26}

\begin{document}

\title{Perceptual Image Compression with Cooperative Cross-Modal Side Information}

\author{
Shi-Yu Qin$^{1,2}$
\quad
Bin Chen$^{ 1,5,6}$\thanks{Corresponding Author} 
\quad
Yu-Jun Huang$^{2}$
\quad
Bao-Yi An$^{3}$
\quad
Tao Dai$^{4}$
\quad
Shu-Tao Xia$^{2}$\\
$^{1}$Harbin Institute of Technology, Shenzhen
\quad\\
$^{2}$Tsinghua Shenzhen International Graduate School, Tsinghua University\quad \\
$^{3}$Huawei Technologies Company Ltd\quad
$^{4}$Shenzhen University \quad
$^{5}$Peng Cheng Laboratory\\
$^{6}$Guangdong Provincial Key Laboratory of Novel Security Intelligence Technologies
\\
{\tt\small 190110427@stu.hit.edu.cn, chenbin2021@hit.edu.cn, huangyj22@mails.tsinghua.edu.cn}
\\
{\tt\small anbaoyi@huawei.com, daitao.edu@gmail.com, xiast@sz.tsinghua.edu.cn}
}

\maketitle

\begin{abstract}
The explosion of data has resulted in more and more associated text being transmitted along with images. 
Inspired by from distributed source coding, many works utilize image side information to enhance image compression. However, existing methods generally do not consider using text as side information to enhance perceptual compression of images, even though the benefits of multimodal synergy have been widely demonstrated in research.
This begs the following question: \textbf{How can we effectively transfer text-level semantic dependencies to help image compression, which is only available to the decoder?} In this work, we propose a novel deep image compression method with text-guided side information to achieve a better rate-perception-distortion tradeoff. Specifically, we employ the CLIP text encoder and an effective Semantic-Spatial Aware block to fuse the text and image features. This is done by predicting a semantic mask to guide the learned text-adaptive affine transformation at the pixel level.
Furthermore, we design a text-conditional generative adversarial networks to improve the perceptual quality of reconstructed images. 
Extensive experiments involving four datasets and ten image quality assessment metrics demonstrate that the proposed approach achieves superior results in terms of rate-perception tradeoff and semantic distortion.
\end{abstract}
\section{Introduction}
\label{sec:intro}

\begin{figure}[t]
    \centering  
    \includegraphics[width=1\linewidth]{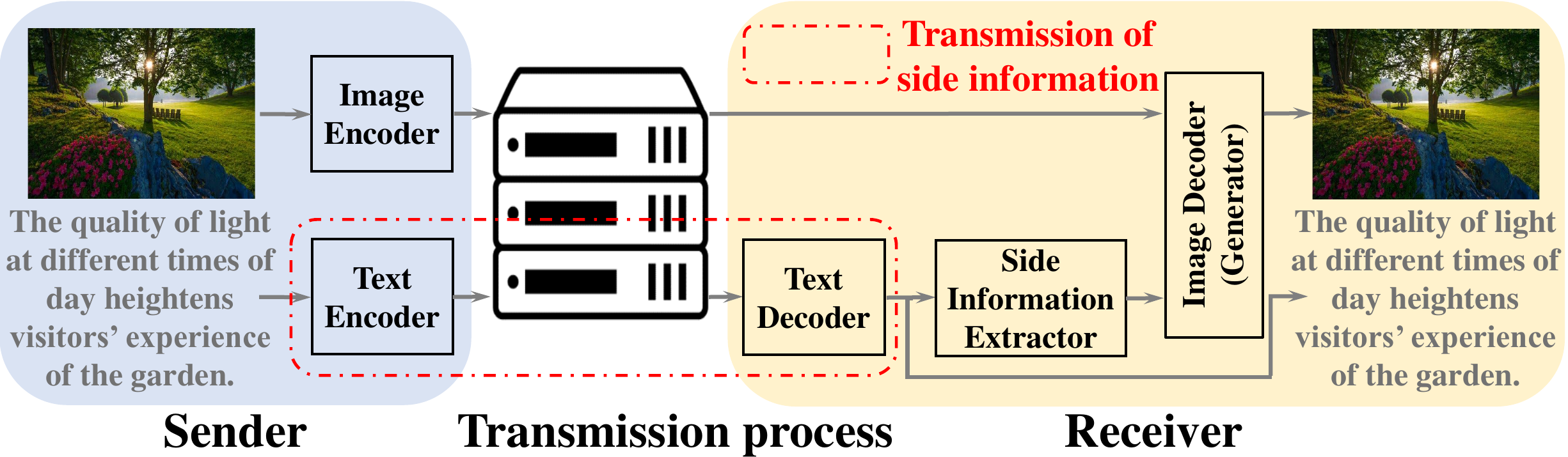} 
    \caption{Scenarios where associated text and images are transmitted together through news media. The text completes the transmission independently and is only involved in the image receiver.} 
    \label{background}
    \vspace{-8pt}
\end{figure}

Over the past few decades, the number of images captured and transmitted via the Internet has exploded, accompanied by associated text. As shown in Figure \ref{background}, news media usually attach pictures, charts and images to report news events, which better illustrate the story. Social media platforms also allow users to post both text messages and images. However, the scheme used in most of these practical scenarios is to encode images and text separately for transmission, which does not take into account the correlation between the two sources. Furthermore, text information is important for image compression as it can provide valuable contextual and descriptive data that complements the visual content of the image. Incorporating text information can aid in better understanding the content of the image, which in turn can help improve compression efficiency. 

Additional textual guidance has been shown to effectively preserve semantic information and enhance perceptual quality (for instance, FID) in generating images for diverse high-level visual tasks~\cite{ramesh2022hierarchical,crowson2022vqgan,nichol2021glide}. In theory, ~\cite{Alpher07} conceptualized 'distortion' as a measure of similarity between image pairs and defined 'perceptual quality' as the disparity between the original image distribution $p_X$ and the reconstructed image distribution $p_{\hat{X}}$, quantified by the dispersion of these distributions. This suggests that better perceptual quality may paradoxically result in increased distortion. Interestingly, recent advancements in image compression reveal that focusing solely on minimizing distortion metrics does not always yield high-quality images. Instead, certain perceptual metrics appear to be more indicative of the quality of image reconstruction. 
However, current approaches~\cite{jiang2023multi} which straightforwardly incorporates text in both the encoding and decoding stages, fails to fully harness the potential of textual information, resulting in a substantial computational burden on front-end resources.

\begin{figure}[t]
    \centering  
    \includegraphics[width=1\linewidth]{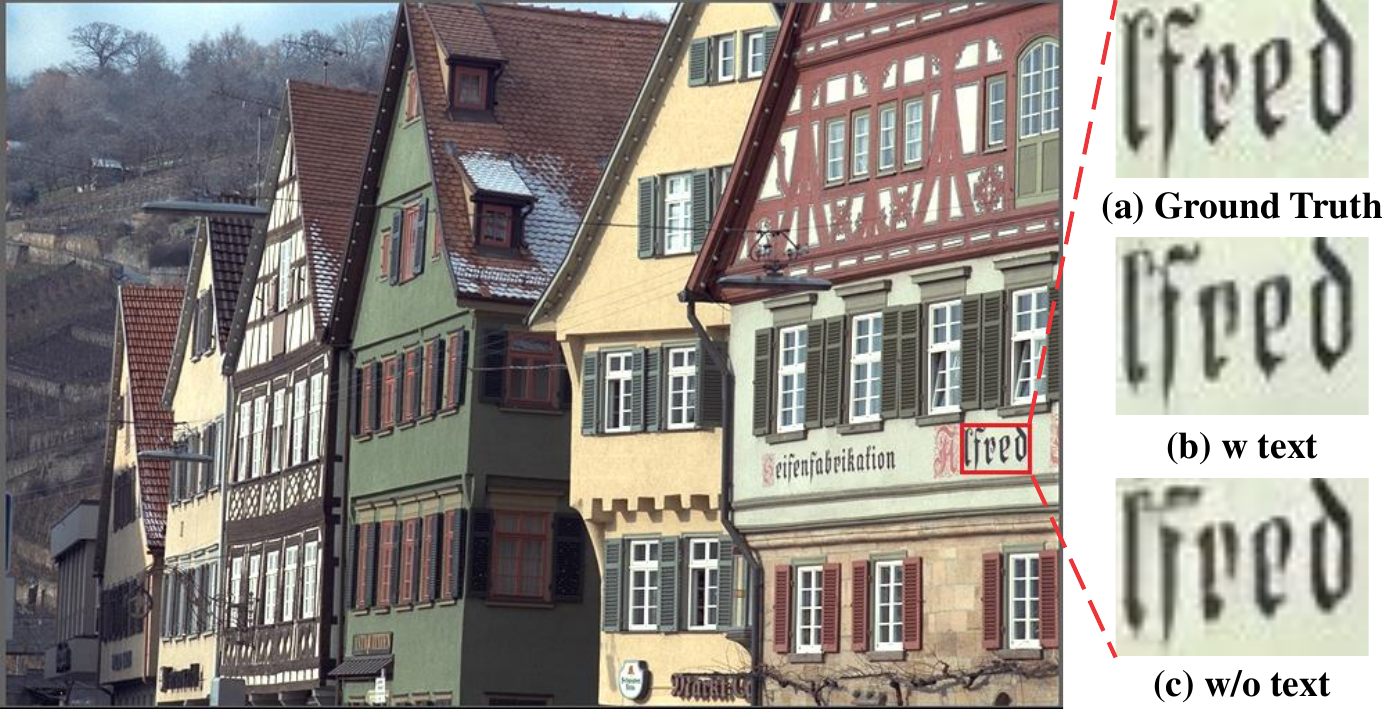}  
    \caption{Visual comparison of ground truth (a) and reconstructed images  (b) (bpp=0.3898, LPIPS=0.05980) and (c) (bpp=0.3934, LPIPS=0.06225) with/without text side information, respectively. The text is \emph{``The building on the right has words inscribed on its facade, adding a unique character to its historic appearance".}} 
    \label{mask}
    \vspace{-8pt}
\end{figure}

Inspired by the distributed source coding (DSC) theory \cite{slepian1973noiseless,wyner1976rate}, DSC theoretically achieves the same compression ratios as the joint encoding scheme, even though the side information is not available to the encoder. Many efforts \cite{Alpher09,Alpher35,Alpher22} have been devoted on employing correlated side information, e.g., image captured from different angles, to improve the image decompression. Plus, they all focus on exploring the abundant texture details in a single modality and none consider the effect of cross-modal side information, which holds more promise and stable semantic information to benefit perceptual DIC. 
Given these promise, one question naturally arises: \emph{How can we best exploit the ability of cross-modal side information in the powerful text description, and effectively adapt it to solve perceptual DIC?} As shown in Figure \ref{mask}, our approach with text enhances the LPIPS from 0.06225 to 0.05991, offering superior visualization and detail recovery. This underscores the significance of text side information in perceptual DIC.

In this paper, we develop a novel DIC method to achieve better perceptual quality and semantic fidelity via cross-modal text side information. 
As shown in Figure \ref{Comparison images}, the text-based side information is only available at the decoder to guide the image decompression. 
Meanwhile, as the demand for data analysis and semantic monitoring has increased in recent years, text-based side information provides an alternative way to compensate for the semantic loss in DIC. 

Specifically, we incorporate the CLIP text encoder\cite{radford2021learning} and Semantic-Spatial Aware (SSA) block \cite{Alpher16} into our framework to exploit text side information. The semantic masks obtained by the SSA module indicate the specific areas where the reconstructed image requires enhancement with textual information at the pixel level. Additionally, to preserve the perceptual quality and semantic relevance of the reconstructed image, we introduce extra textual information and the corresponding discriminative loss function in the discriminator. We evaluate our methodology using metrics from ten different perspectives and demonstrate the improvement in perceptual quality and compensation for semantic loss in DIC. Furthermore, we showcase the resilience and versatility of text as supplementary information by applying different text inputs to the same image and various compression models.

To summarize, our contributions are as follows:
\begin{itemize}
 \item We explore the role of text as a type of cross-modal side information to guide image perceptual compression.
\item We integrate CLIP text encoder as side information extractor and the Semantic-Spatial Aware block to exploit text side information, which helps to enhance the image decompression with textual information at the pixel level.
 \item We quantitatively evaluate our method under ten metrics including distortion metrics, perceptual metrics, semantic relevance metrics, and semantic precision to demonstrate that the proposed method achieves superior results.
\end{itemize}

\section{Related Work}
\label{sec:Related Work}

\noindent\textbf{Deep Image Compression.} Deep Image Compression, which uses Deep Neural Networks (DNN) for image compression, has achieved great success with promising results. ~\cite{Alpher17} first proposed an end-to-end image compression model based on rate-distortion optimization. ~\cite{Alpher18} further proposed a hyperprior model based on this, which captures the redundancy in the hidden feature space and better estimates the code rate. ~\cite{Alpher19} introduced GAN at the decoding side to reduce artifacts and improve the visual quality. ~\cite{mentzer2020high} propose a generative compression method to achieve high fidelity reconstructions. 
In our compression framework, we take advantage of the advanced transform (network structures), quantization and entropy models in the existing learning-based methods for lossy image compression. 

\begin{figure}[h]
    \centering  
    \includegraphics[width=1\linewidth]{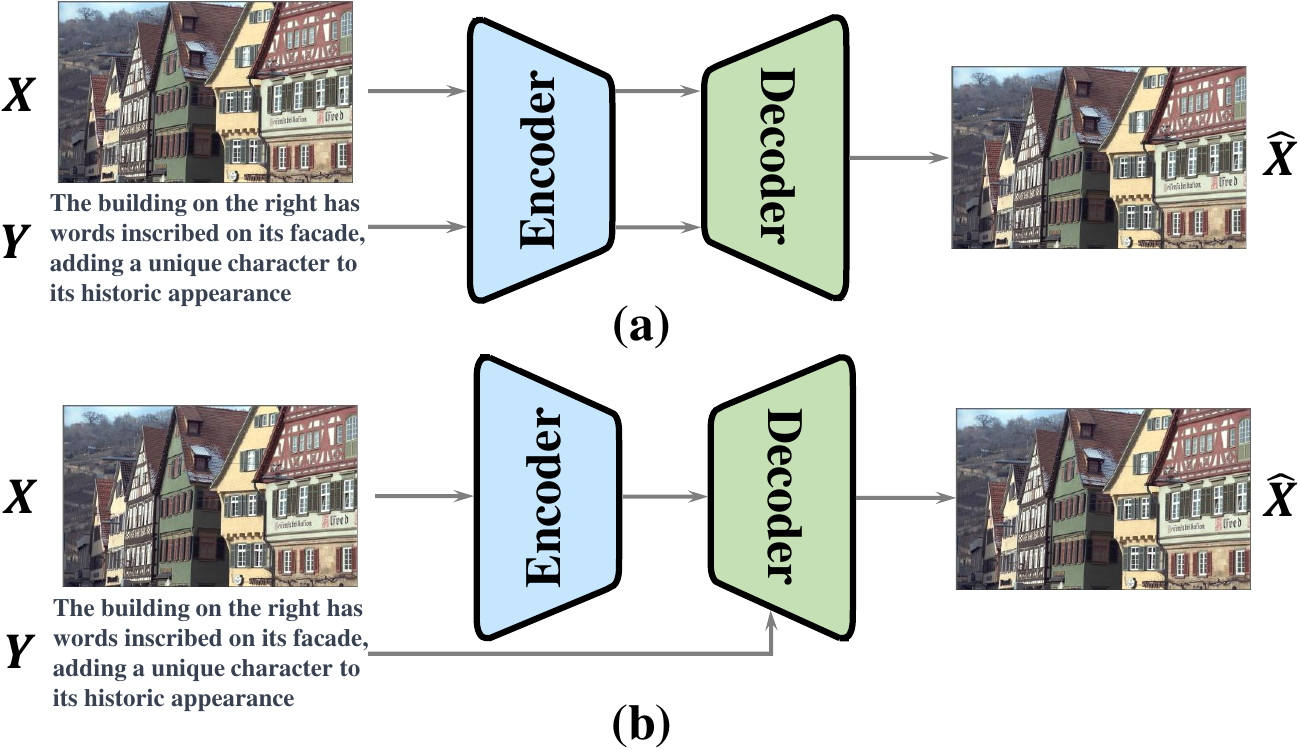} 
    \caption{The comparison of (a) joint image-text coding \cite{jiang2023multi} and our proposed distributed coding with text side information (b).} 
    \label{Comparison images}
\end{figure}

\begin{figure*}[t]
    \centering  
    \includegraphics[width=1\linewidth]{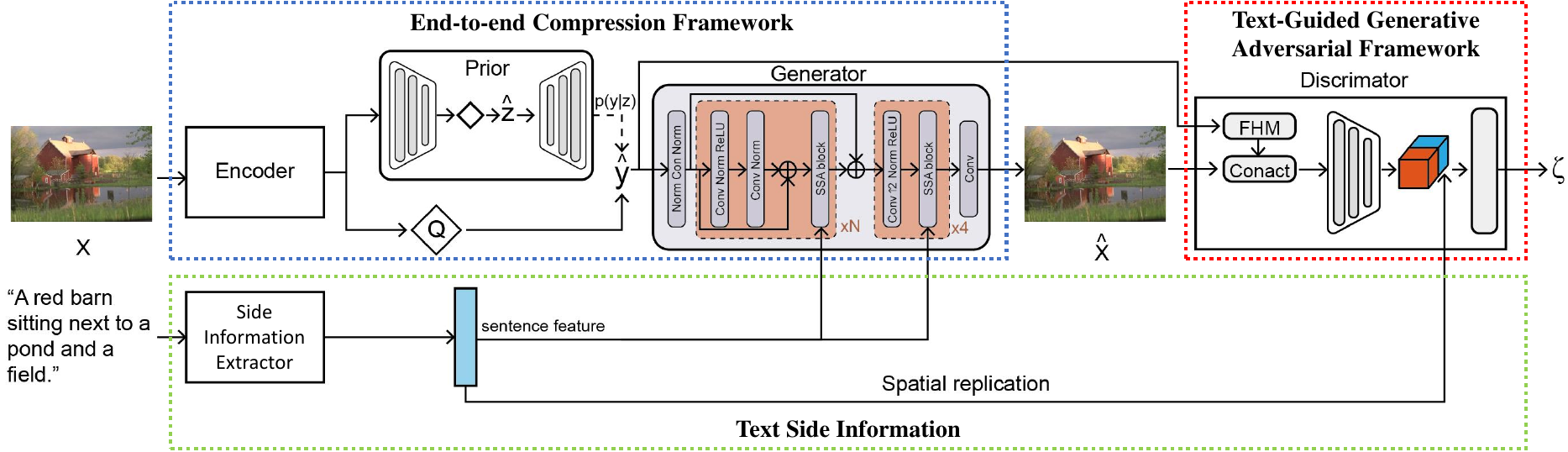}  
    \caption{Overview of our model.} 
    \label{backbone_main}
\end{figure*}

\noindent\textbf{Learned Distributed Source Coding.} David Slepian and Jack K. Wolf proposed the Slepian-Wolf bound in 1973, which proved that compression of two information-related sources that do not communicate with each other could achieve compression efficiency with mutual communication, marking the birth of distributed source coding. Similar results were also obtained by Wyner and Ziv \cite{wyner1976rate} with regard to lossy coding of joint sources with side information. Recent works employed deep learning to implement distributed source coding. ~\cite{Alpher09} proposed that the aligned side information image can be used to better reconstruct the main image. \cite{Alpher22} proposed that patch matching in the multi-scale feature domain performs better than in the image domain. In fact, the most closely related research to ours is \cite{jiang2023multi} that on adopting text description in different components, i.e., the encoder, entropy coding and the decoder of the image codec (first row of Figure \ref{Comparison images}). But we consider a different scenario where the text side information is only available at the decoder side (second row of Figure \ref{Comparison images}), because DSC \cite{cover1999elements} proves their equivalent compression efficiency in theory, and mitigates the computational complexity at the encoder in practice. However, there still lack a practical method to use more concise text side information to aid perceptual image compression.

\section{Proposed Method}

\subsection{Overview}

Firstly, we present an overview of our proposed method. Drawing from the classical distributed source coding with side information in information theory, we posit that the cross-modal side information is only accessible to the decoder. However, unlike the current learned distributed source coding methods within a single modality, e.g., solely in the image domain as depicted in Figure \ref{Comparison images} (a), we explore the efficacy of cross-modal side information as illustrated in Figure \ref{Comparison images} (b). This involves the use of strongly correlated text inputted into the decoder to aid in decompression. As illustrated in Figure \ref{backbone_main}, our architecture is composed of five elements: an image encoder $E$ that extracts compressible features, an entropy prior model $P$, an image decoder (generator) $G$ that recovers compressed images with the help of side information, a side information extractor to extract text features, and a discriminator $D$ that engages in a zero-sum game within an adversarial framework to enhance perceptual fidelity. Once the training is completed, only the generator is utilized for decompression.

\begin{figure}[t]
    \centering  
    \includegraphics[width=1\linewidth]{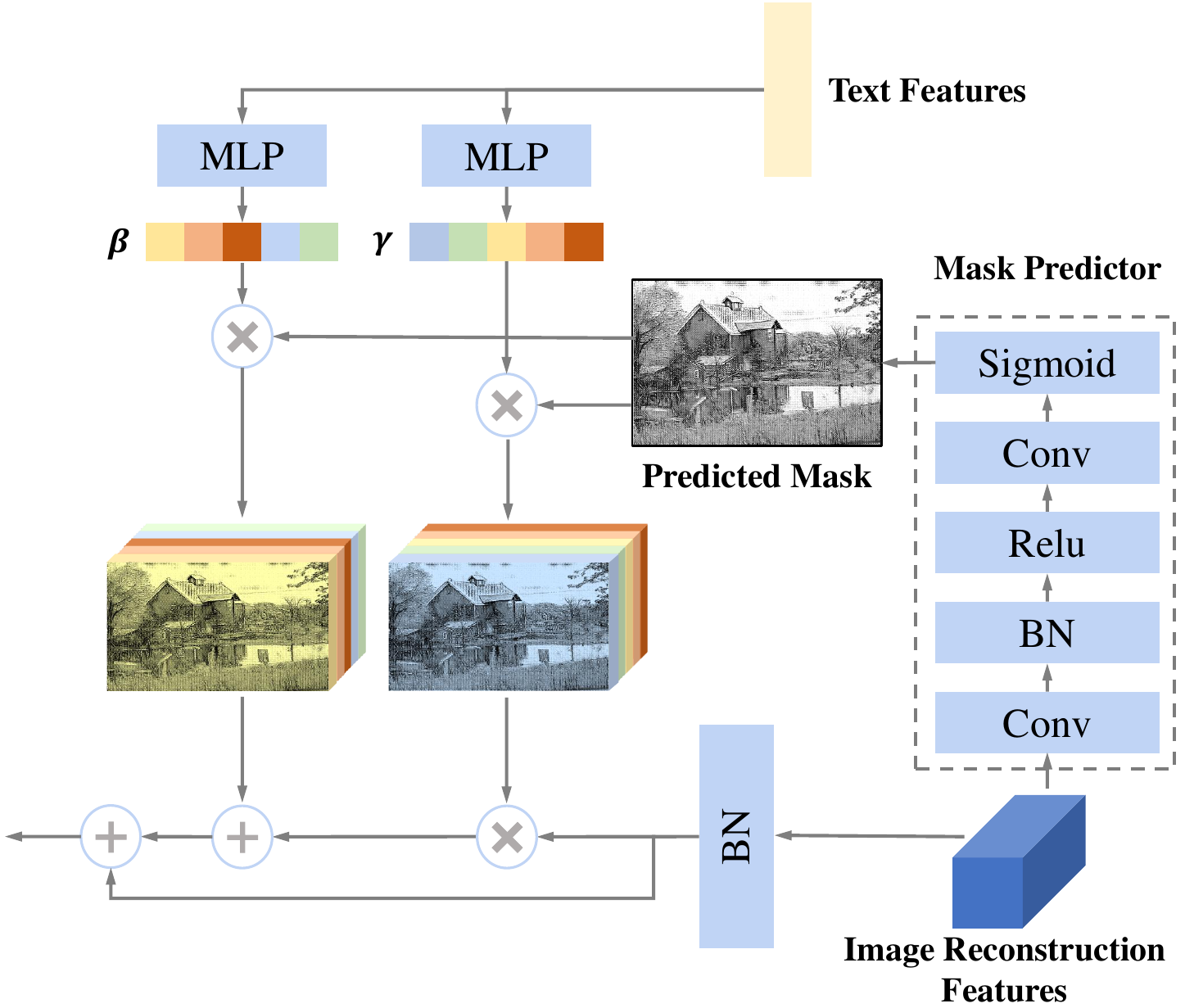}  
    \caption{Structure of the SSA block. $\oplus$/$\otimes$ denotes elementwise addition/multiplication.} 
    \label{SSA block}
    \vspace{-16pt}
\end{figure}

\begin{figure*}[t]
    \centering  
    \includegraphics[width=1\linewidth]{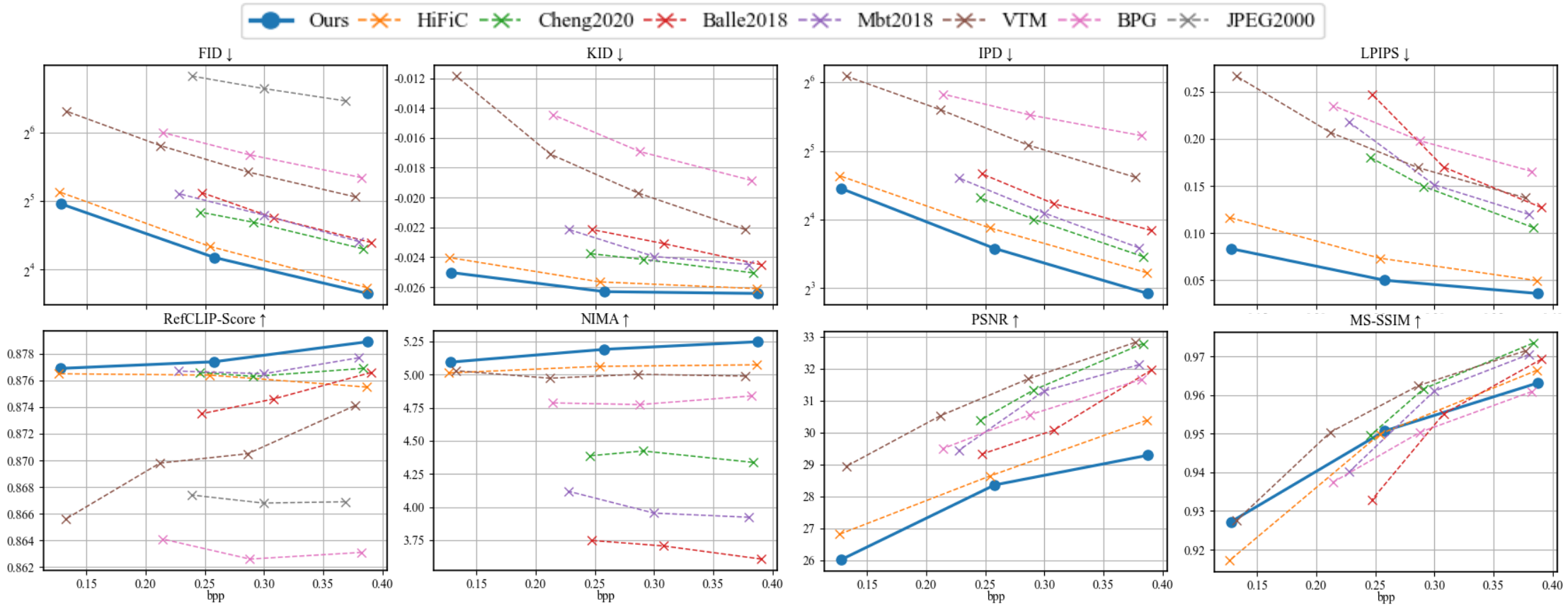} 
    \caption{Different metrics on Kodak~\cite{franzen1999kodak}. Arrows in the plot titles indicate whether high (↑) or low (↓) values indicate a better score.} 
    \label{Kodak}
\end{figure*}

Specifically, our encoder network $E(\cdot,\theta)$ takes an uncompressed image $\mathbf{x}\in\mathbb{R}^{H \times W \times 3}$ and output compressible feature $\mathbf{y}\in\mathbb{R}^{{H^{'}}\times{W^{'}}\times C}$, which is immediately fed to the quantizer $Q$ to get the quantized features $\mathbf{\hat{y}}$. The priori networks $P(\cdot,\psi)$ learn the distribution of compressed features $p(\mathbf{y}|\mathbf{z})$ to predict code rates. The generator network $G(\cdot,\gamma)$ takes the quantized features $\mathbf{\hat{y}}$ and the text features $\mathbf{I}$ extracted by the side information extractor as input and outputs the reconstructed image $\mathbf{\hat{x}}\in\mathbb{R}^{H \times W \times 3}$. In the training phase, we introduce a discriminator $D(\cdot)$ with side information to determine whether the image and text match. After the text features are spatially replicated and fused with the image features in spatial dimensions, the discriminator outputs the final cross-entropy loss $L$. We continue to include the hidden feature $\mathbf{y}$ as an additional condition in HiFiC~\cite{mentzer2020high} and refer to the feature alignment network in the discriminator as the feature harmonization module (FHM).

\subsection{Joint Training with Text Side Information}
\noindent\textbf{Side Information Extractor.} We employ the text encoder from the CLIP model \cite{radford2021learning}, which was pre-trained along with the corresponding image encoder, to convert the input text information into a text vector $\mathbf{\bar{v}} \in \mathbb{R}^{512}$. Its parameters are kept frozen during the main model training.

\noindent\textbf{Generator (Decoder) with Text Side Information.} According to the distributed source coding theory with side information, additional side information needs to be added to the decoder side to guide the image decompression. Therefore, we design a novel image decoder, which introduces the semantic-spatial aware block (as shown in Figure \ref{SSA block}) of SSA-GAN~\cite{Alpher16} into the decoder. The original decoder consists of multiple convolutional layers and upsampling layers. Its input is the hidden feature $\mathbf{\hat{y}} \in \mathbb{R}^{{C} \times {16} \times {16}}$ obtained from entropy decoding, and the output is a color RGB image $\mathbf{\hat{x}} \in \mathbb{R}^{{3}  \times {256} \times {256}}$ after convolution and four times of upsampling. So we divide the decoding process into convolutional residual block and convolutional upsampling block according to the network structure and introduce SSA block at the end of each part.

Each SSA block consists of a semantic mask predictor, a semantic-spatial condition batch normalization block with a residual connection. It takes the encoded text features $\mathbf{\bar{v}}$ and the reconstruction features of the previous stage $\mathbf{y}_i \in \mathbb{R}^{c_i \times h_i \times w_i}$ as input, where $c_i$, $h_i$, $w_i$ are the number of channels, height and weight of the reconstructed features, and outputs the reconstructed features that are better integrated with the text features. For semantic mask predictor, it takes the image reconstruction features from the previous stage as input and predicts a mask map $\mathbf{m}_i \in \mathbb{R}^{h_i \times w_i}$, whose component takes value between [0,1]. Note that each value $\mathbf{m}_{i}(h, w)$ determines to what extent the following transformations should be manipulated at that position. The prediction mask is based on the reconstructed features of the previous stage, so it can visually reflect which parts still need to be enhanced with textual information to guide the decompressed image and maintain their semantic consistency. Following this, the affine transformation parameters $\gamma_c$ and $\beta_c$ are learned from the textual features $\mathbf{\bar{v}}$ using two multilayer perceptrons:
\begin{equation} 
\gamma_c=P_\gamma(\mathbf{\bar{v}}),\quad\beta_c=P_\beta(\mathbf{\bar{v}}),
\end{equation}

We then perform a spatial affine transformation of the normalized reconstructed features by the prediction mask and affine transformation parameters:
\begin{equation} 
	\mathbf{y}_{i+1}=m_{i,(h, w)}\left(\gamma_c(\mathbf{\bar{v}}) \mathbf{\hat{y}}_i+\beta_c(\mathbf{\bar{v}})\right),
\end{equation}
where $\mathbf{\hat{y}}_i$ is the normalized reconstructed features $\mathbf{y}_i$.

\begin{figure*}[t]
    \centering  
    \includegraphics[width=1\linewidth]{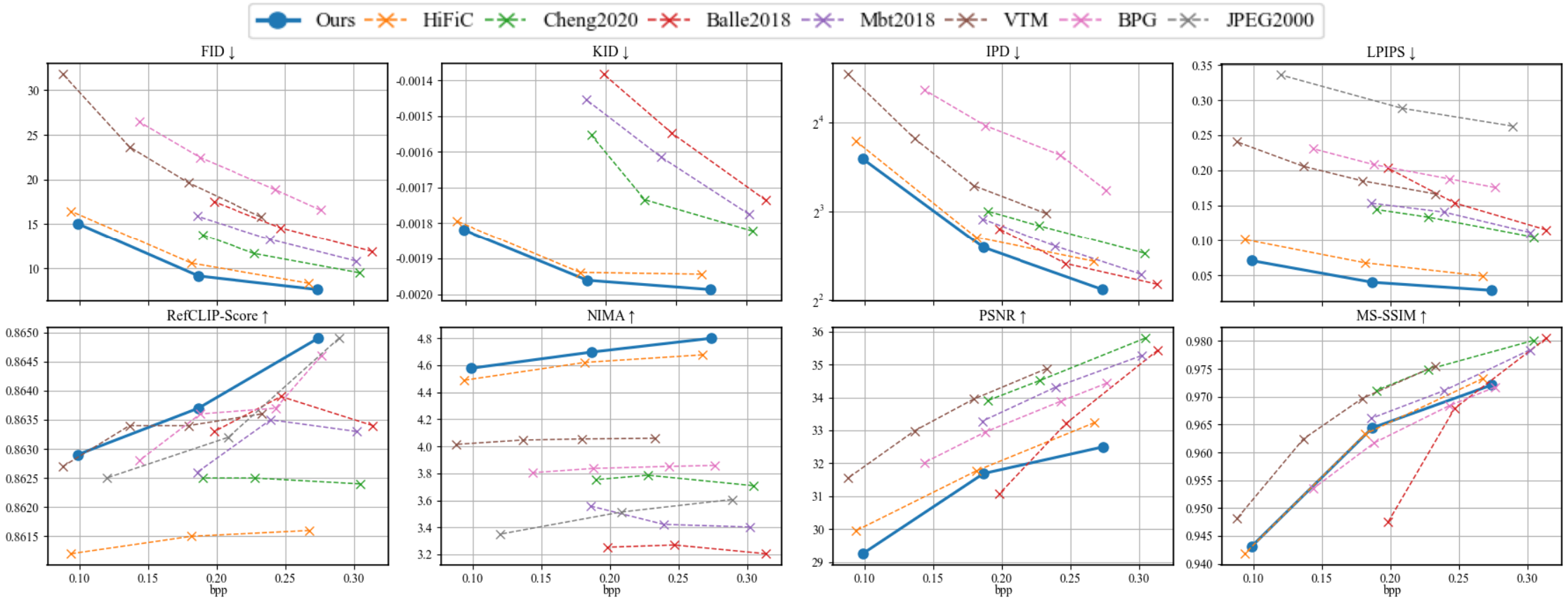}  
    \caption{Different metrics on CLIC2020~\cite{toderici2020clic}. Arrows in the plot titles signify whether high (↑) or low (↓) values represent a better score.} 
    \label{CLIC2020}
\end{figure*}

\subsection{Objective Functions}
\noindent\textbf{Generator Objective.} 
The main task of image compression is a rate-distortion optimization. Therefore, 
the total loss for the Encoder, Prior and Generator is composed of distortion, code rate and an adversarial loss:
\begin{equation}
	\mathcal{L}_{E G P}=\mathbb{E}_{\mathbf{x \sim p_X}}\left[\lambda(r(\mathbf{\hat{y}})+r(\mathbf{\hat{z}}))+d\left(\mathbf{x}, \mathbf{\hat{x}}\right)+\beta \mathcal{L}_{a d v}^G\right],
\end{equation}
where $\mathcal{L}_{a d v}^G$ denotes the generative adversarial loss for the generator:
\begin{equation}
\mathcal{L}_{a d v}^G=-E_{\mathbf{x} \sim \mathbf{p}_{\mathbf{G}}}[D(\hat{\mathbf{x}}, \hat{\mathbf{y}}, \mathbf{s})].
\end{equation}

To measure the bit-rate of a compressed image, we define $r(\mathbf{\hat{y}})$ and $r(\mathbf{\hat{z}})$ to be the rate terms, calculated by $\mathbb{E}\left[-\log _2\left(p_{\hat{\boldsymbol{\mathbf{y}}} \mid \hat{\boldsymbol{\mathbf{z}}}}(\hat{\boldsymbol{\mathbf{\mathbf{y}}}} \mid \hat{\boldsymbol{\mathbf{\mathbf{z}}}})\right)\right]$ and $\mathbb{E}\left[-\log _2\left(p_{\dot{\boldsymbol{\mathbf{z}}} \mid \boldsymbol{\psi}}(\hat{\boldsymbol{\mathbf{z}}} \mid \boldsymbol{\psi})\right)\right]$, where $\psi$ is given by the entropy models for $\mathbf{\hat{z}}$. We then introduce a distortion loss $d\left(\mathbf{x}, \mathbf{\hat{x}}\right)$ consisting of MSE and the perceptual metric LPIPS~\cite{Alpher29} as follows: 
\begin{equation}
	d=k_Md_{MSE} + k_P d_{LPIPS}.
\end{equation}

In the above equation, $\lambda$, $\beta$, $k_M$, and $k_P$ are the hyperparameters of the control loss, respectively.

\noindent\textbf{Discriminator Objective.}
We introduce text side information in the discriminator to improve the semantic relevance of the image text, which not only has to discriminate whether the compressed image is real enough but also whether the input image matches the input text. Therefore, the adversarial loss of the discriminator consists of three parts:
\begin{eqnarray}
\mathcal{L}_D&=&\mathbb{E}_{x \sim p_G}\left[-\log \left(1-D\left(\hat{\mathbf{x}}, \hat{\mathbf{y}}, \mathbf{s}\right)\right)\right] \\
\nonumber&+&\mathbb{E}_{x \sim p_X}[-\log (D(\mathbf{x}, \hat{\mathbf{y}}, \mathbf{s}))] \\
&+& \mathbb{E}_{x \sim p_X}\left[-\log \left(1-D\left(\mathbf{x}, \hat{\mathbf{y}}, \hat{\mathbf{s}}\right)\right)\right]\nonumber,
\end{eqnarray}
where $\mathbf{s}$ is the given text description while $\mathbf{\hat{s}}$ is a mismatched
text description. $\mathbf{x}$ is the real image described by $\mathbf{s}$, $\mathbf{\hat{x}}$ is the reconstructed images. $\mathbf{\hat{y}}$ is the feature representation obtained from the encoder. $D(\cdot)$ represents the discriminator's decision on whether the input image and text match.
\section{Experiments}

\begin{figure*}[t]
    \centering  
    \includegraphics[width=1\linewidth]{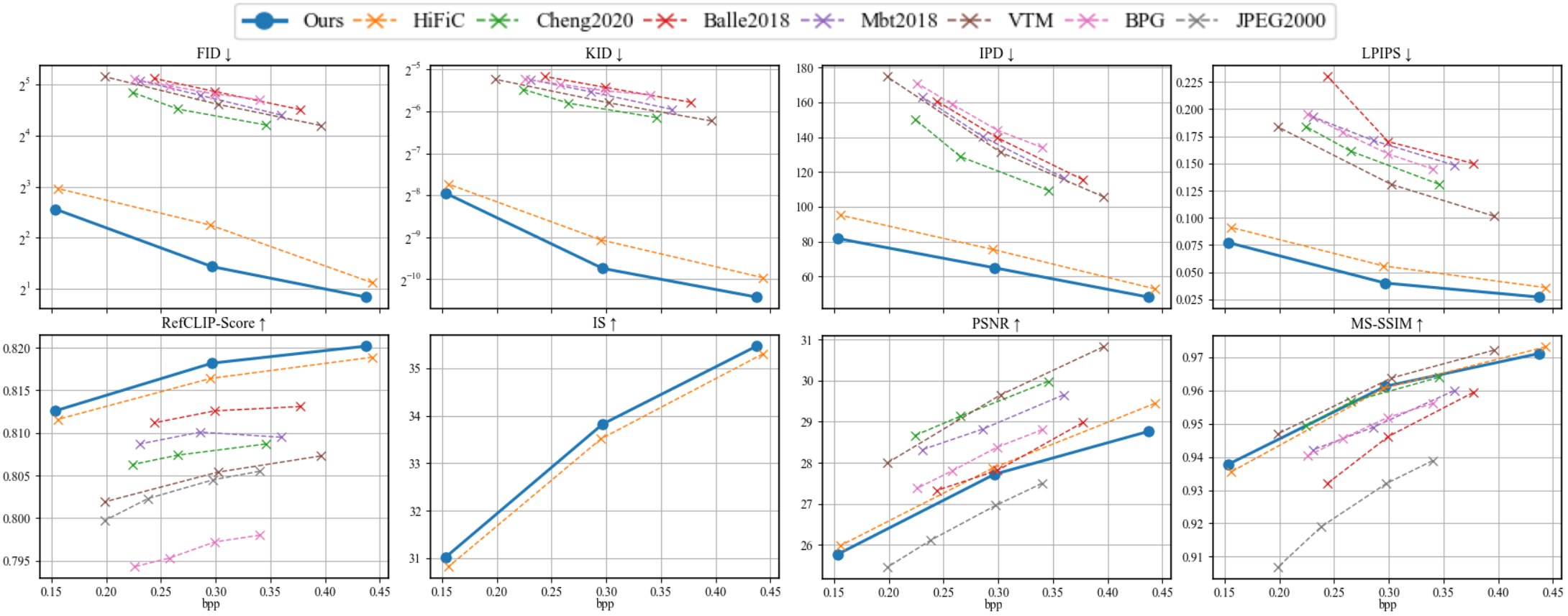}  
    \caption{Different metrics on MS COCO~\cite{Alpher27}. Arrows in the plot titles indicate whether high (↑) or low (↓) values indicate a better score.} 
    \label{MSCOCO}
\end{figure*}

\subsection{Experimental Setup}
\noindent\textbf{Datasets.} We conduct experiments to show qualitative and quantitative results on both MS COCO~\cite{Alpher27}, Kodak~\cite{franzen1999kodak}, CLIC2020~\cite{toderici2020clic} and CUB~\cite{wah2011caltech}. The MS COCO dataset ~\cite{Alpher27}contains 82,783 training images and 40,504 test images, each with five descriptive texts. The Kodak dataset~\cite{franzen1999kodak} and the CLIC2020 dataset~\cite{toderici2020clic} have 24 and 428 images, respectively. CUB~\cite{wah2011caltech} consists of 200 species of bird, with a total of 11,788 images including 8,855 images for training and 2,933 images for testing. We used BLIP2~\cite{li2023blip} to extract one sentence of text per image for the CLIC2020 dataset~\cite{toderici2020clic} and GPT4 to extract five sentences of text per image for the Kodak dataset~\cite{franzen1999kodak} while train our model using the training set of the MS COCO dataset~\cite{Alpher27}. For each image, we did not crop considering that different parts correspond to different semantic information.

\noindent\textbf{Evaluation Metric.} 
In order to provide a comprehensive assessment of the proposed approach, ten evaluation metrics are adopted. These include perceptual quality evaluation metrics LPIPS~\cite{Alpher29}, image fidelity metrics FID~\cite{Alpher30}, KID~\cite{binkowski2018demystifying} and IPD~\cite{Alpher21}, image text relevance metrics  RefCLIP-Score~\cite{Alpher31}, image diversity metrics IS~\cite{Alpher29}, human assessment metrics NIMA~\cite{talebi2018nima}, distortion metrics, and multi-categorization task~\cite{Alpher32} accuracy metrics. At the same time, these metrics review each method from the perspective of instances(LPIPS, IPD, PSNR, etc.) and collections(FID, KID, etc.), respectively. 

\noindent\textbf{Implementation details.}
Our model is implemented using Pytorch. We employed the most cutting-edge high-fidelity image compression network~\cite{mentzer2020high} as the foundation. The experimental setup we used is identical to that of HiFiC~\cite{mentzer2020high}. The entire training process is split into two stages. In the first stage, we solely utilize the rate-distortion loss function to train the encoder $E$, the prior model $P$, and the generator (decoder) $G$, with a batch size of 8. In the second stage, we concurrently update the discriminator and the discriminative loss, with a batch size of 16. We employ the Adam optimizer in both stages. This configuration is designed to enhance the stability of model training. The hyper-parameters $\beta$ = 0.15, $k_M$ = 1, $k_P$ = 0.075·$2^{-5}$, and $N = 9$. Subsequently, we trained three models with bpp approximating ${0.15, 0.3, 0.45}$ respectively. The model is trained on the MS COCO dataset~\cite{Alpher27} for 200 epochs at each stage.

\begin{figure}[t]
    \centering  
    \includegraphics[width=0.7\linewidth]{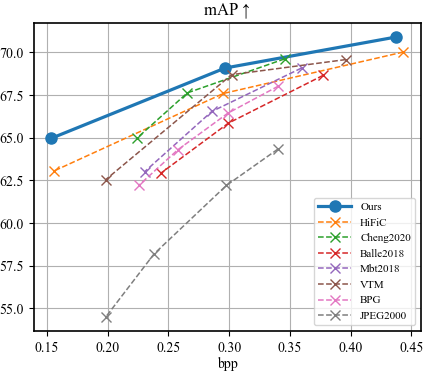} 
    \caption{Rate-mAP curve on MS COCO~\cite{Alpher27}.} 
    \label{mAP}
    \vspace{-8pt}
\end{figure}

\subsection{Comparison with Other Methods}
\noindent\textbf{Rate-Distortion-Perception Performance.}  In Figure ~\ref{Kodak}, Figure ~\ref{CLIC2020}, and Figure ~\ref{MSCOCO}, we present rate-distortion, rate-perception, and rate-semantic relevance curves evaluated using ten metrics on MS COCO~\cite{Alpher27}, Kodak~\cite{franzen1999kodak}, and CLIC2020~\cite{toderici2020clic}. Our model is compared with TGIC~\cite{jiang2023multi}, HiFiC~\cite{mentzer2020high}, Cheng2020~\cite{Alpher20}, Belle2018~\cite{Alpher18}, Mbt2018~\cite{minnen2018joint}, VTM, BPG~\cite{bpg}, and JPEG-2000~\cite{Alpher01}. Since ~\cite{jiang2023multi} have not publicly released the source code and the reproduction results are subpar, we independently compared our method with theirs on the CUB~\cite{wah2011caltech}, utilizing the original performance curves. The detailed results can be found in the appendix. Our model significantly outperforms other baseline models in terms of perceptual quality, image fidelity, and semantic relevance metrics, and also aligns with expectations by reducing distortion metrics.

To verify the practical usefulness of the improved perceptual quality and semantic relevance, we evaluate the accuracy (mAP)-bpp tradeoff on a multi-label classification task~\cite{Alpher32}. As shown in Figure~\ref{mAP}, our model also outperforms other methods in classification accuracy. This suggests that the introduction of text supplements the loss of semantic accuracy incurred during compression thereby making the reconstructed image more straightforward to understand by machine recognition.

\begin{figure*}[t]
    \centering  
    \includegraphics[width=1\linewidth]{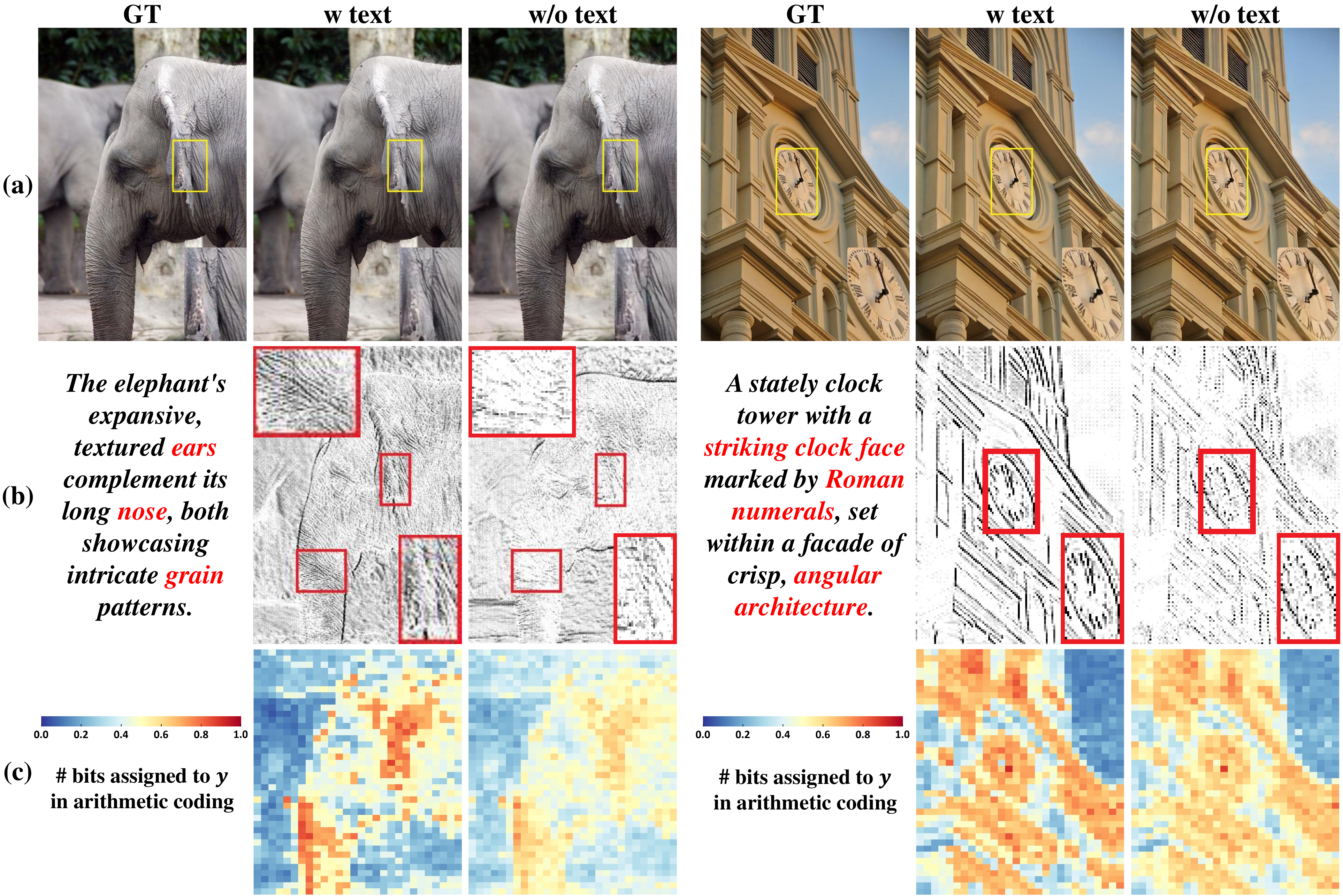}  
    \caption{Visualization of decoded images with (Left, bpp=0.2549, LPIPS=0.05557; Right, bpp=0.3263, LPIPS=0.02420) and without (Left, bpp=0.2671, LPIPS=0.06986; Right, bpp=0.3245, LPIPS=0.03363) text (first row), text side information and corresponding prediction masks (second row) and bit allocation maps of latent $ \hat{\textbf{y}} $ (third row). The values of bit allocation maps denote the average negative log likelihood of each element in $ \hat{\textbf{y}} $ across all channels.} 
    \vspace{-8pt}
    \label{bit_map}
\end{figure*}

\begin{figure*}[t]
    \centering  
    \includegraphics[width=1\linewidth]{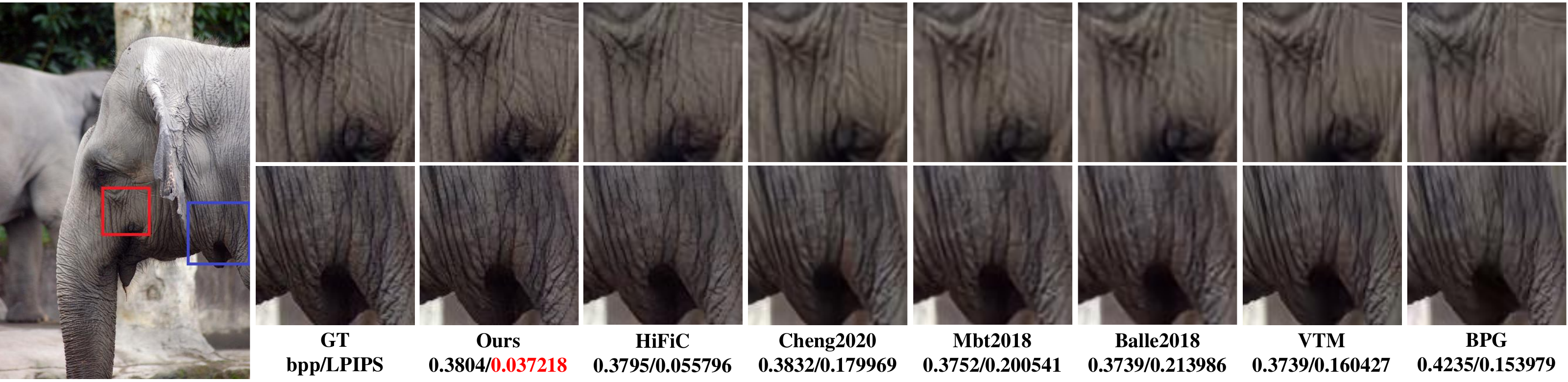}  
    \caption{Visualization of different image compression codecs.} 
    \label{Visualization}
\end{figure*}

\noindent\textbf{Qualitative Results.} Furthermore, we showcase visual comparisons of various methods, as depicted in Figure~\ref{Visualization}. It is clear that our method restores detailed textures (for instance, the wrinkles in an elephant's skin) more effectively, without blocky blurring, and with superior perceptual quality. Our model exemplifies the efficiency of compression and the precision of restoration.

\subsection{Contributions of Text Side Information}
\noindent\textbf{Bit allocation and prediction mask based on textual semantics.} Figure \ref{bit_map} illustrates the role of text side information in compression. The text (first column of (b)) emphasizes local features (e.g., nose and ears) as well as visual effects (e.g., texture features) in the image to be compressed.The prediction masks generated by the SSA module of the model using the text based on this semantic information assign more weight to the texture part of the ears and nose, whereas the prediction masks generated by the model without the guidance of the text do not have this feature (last two columns of (b)). The prediction mask gradually guides the entropy model in the gradient update to assign more bits to the parts with more weight, and (c) demonstrates that the model based on textual semantics assigns more bits to the ears, nose, and texture-dense parts, and fewer bits to the parts that are not mentioned in the text (e.g., the background parts). In contrast, the ablative model allocates more bits to the background part (lighter blue) and fewer bits to the parts emphasized by the text (lighter red).

\noindent\textbf{Ablation Study.}
To demonstrate the role of text, we trained different variants of the proposed method including: removing text at the generator side, removing text at the discriminator side and removing the overall text, where removing text refers to the use of zero features instead of text features. In the inference phase, the model that removes only the discriminator-side text has normal text input, while the models that remove only the generator-side text and remove all text do not have text input and use zero features instead.

\begin{table}[h]
\scriptsize
   \centering
   \caption{Partial BD-Rate results for different variants on Kodak~\cite{franzen1999kodak}. The baseline is the full model.}
       \label{BD_rate}
       \footnotesize
       \setlength{\tabcolsep}{1.5mm}{
        \begin{tabular}{c|c|c|c|c}
        \hline
        $\textbf{Variant}$&$\textbf{BD-Rate}_L$&$\textbf{BD-Rate}_F$&$\textbf{BD-Rate}_K$&$\textbf{BD-Rate}_I$\\ 
        \hline
            w/o D text&$4.02\%$&$5.22\%$&$16.40\%$&$9.41\%$ \\
            w/o G text&$5.89\%$&$9.77\%$&$19.50\%$&$15.69\%$ \\
            w/o text&$11.51\%$&$18.70\%$&$29.59\%$&$25.81\%$ \\
        \hline
        \hline
        \end{tabular}
    }
    \vspace{-8pt}
\end{table}

We provide visual comparisons (Figure \ref{mask} and Figure \ref{bit_map}) and quantitative comparisons of metrics for ablation experiments. Table \ref{BD_rate} shows the BD-rate of the different variants of the model on the Kodak dataset~\cite{franzen1999kodak} for partial metrics(LPIPS, FID, KID, IPD), where the values indicate the percentage increase in rate of the various variants compared to the full model, with larger values indicating more bits required compared to the full model. The models that use text only at the decoder side and only at the discriminator side both consume fewer bits than the model that removes the overall text. In addition, using text on the decoder-only side consumes fewer bits than using text on the discriminator-only side, which is as expected. The text on the decoder side is more important than the text on the discriminator side, as it is directly involved in the decoding process and directs the image as to where exactly semantic enhancement is needed, while the text on the discriminator side determines whether the text is adequately represented in the image and whether the image and text match.

\subsection{Stability Evaluation}
Considering that the same image has different descriptive texts, we designed robustness experiments to demonstrate that the semantic guidance can be provided stably even with different texts. In the previous inference process, each image in Kodak~\cite{franzen1999kodak} corresponds to five text descriptions with similar semantics, from which we randomly select one as the side information input. To prove that text as side information has stronger robustness, we fixed the number of texts per input and tested the performance metrics using the first to fifth sentences and the wrong sentences as side information in turn, partial results are shown in Table \ref{Robustness_p}. 

\begin{table}[h]
\scriptsize
   \centering
   \caption{Partial performance with different texts on Kodak~\cite{franzen1999kodak}. }
   \label{Robustness_p}
   \footnotesize
    \begin{tabular}{c|c|c|c|c}
    \hline
    \textbf{Sentence}&\textbf{Rate(bpp)}&\textbf{PSNR(dB)↑}&\textbf{LPIPS↓}&\textbf{FID↓}\\ 
    \hline
        1&$0.2579$&$28.3702$&$0.050016$&$18.0892$ \\
        2&$0.2579$&$28.3721$&$0.050010$&$18.0886$ \\
        3&$0.2579$&$28.3699$&$0.050006$&$18.0899$ \\
        4&$0.2579$&$28.3714$&$0.050021$&$18.0884$\\
        5&$0.2579$&$28.3709$&$0.050001$&$18.0901$ \\
        Mismatch&$0.2579$&$28.1968$&$0.107298$&$31.8439$ \\
    \hline
    \hline
    \end{tabular}
\end{table}

Feeding a different text during the inference stage does not impact the bit rate of the compressed image, as the text is only for the decoding process, the entropy coding of the image's hidden features has already been completed prior to this. The text plays a role in the reconstruction of the hidden features obtained by entropy decoding through the decoder. It's noteworthy that the performance of each model is either very close when using different texts, demonstrating the robustness of the text side information. When mismatched text is used, the model's performance declines, with a much larger decrease in perceptual metrics than in distortion metrics. This happens because the mismatched text contains semantic information that is unrelated to the image and cannot effectively guide the image reconstruction process.

\subsection{Variant Backbones}
To integrate phenotypic text-side information that is not solely applicable to a specific deep compression model, thereby ensuring high generalizability, we evaluated our proposed method on Cheng2020~\cite{Alpher20}. We segmented the decoding process into four stages, corresponding to the decoder's four up-sampling procedures, and incorporated the SSA module after each stage. We employed a discriminator with a similar network structure. The comprehensive structure of this network is detailed in the Appendix.

\begin{figure}[t]
    \centering  
    \includegraphics[width=1\linewidth]{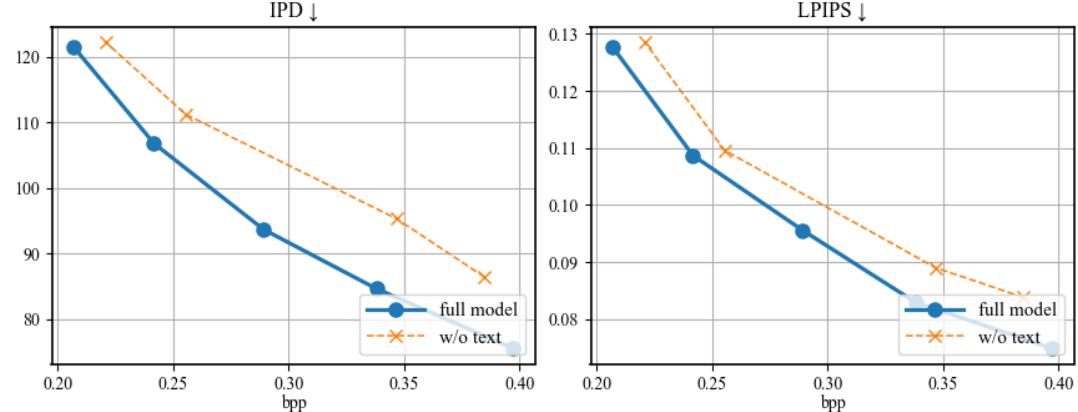}  
    \caption{Partial metrics on MS COCO~\cite{Alpher27} for Cheng2020's approach~\cite{Alpher20} to using text side information.} 
    \label{cheng2020_four_metric}
    \vspace{-8pt}
\end{figure}

Figure \ref{cheng2020_four_metric} illustrates the performance disparity between the models with/without the text side information. It can be seen that including text side information significantly improves the fidelity, perceptual quality, semantic relevance, and classification accuracy of the reconstructed images. This suggests that text side information can enhance the performance of perceptual image compression models as well as the effectiveness of deep compression models focused on signal fidelity.
\section{Conclusion}
This paper systematically investigates whether cross-modal side information is readily applicable for distributed image compression to improve the perceptual quality of reconstructed images and reduce semantic loss. 
Our extensive experiments demonstrate that textual side information enhances the reconstructed image's perceptual quality and semantic relevance. An interesting future direction is to explore different types of additional information for image compression, such as incorporating acoustical signals.

\bibliographystyle{ieee_fullname}
\bibliography{egbib}
\clearpage
\addcontentsline{toc}{section}{Appendix}
\section*{Appendix}
\appendix
\renewcommand{\thetable}{A\arabic{table}}
\renewcommand{\thefigure}{A\arabic{figure}}

\end{document}